\documentclass[conference]{IEEEtran}
\IEEEoverridecommandlockouts
\usepackage{cite}
\usepackage{amsmath,amssymb,amsfonts}
\usepackage{algorithm, algorithmic}
\usepackage{graphicx}
\usepackage{textcomp}
\usepackage{xcolor}
\usepackage{dsfont}
\usepackage{comment}
\usepackage{hyperref}
\usepackage{tikz}
\usetikzlibrary{positioning}
\tikzstyle{obs} = [circle, draw=black, fill=yellow!30, thick, minimum size=7mm]
\tikzstyle{state} = [circle, draw=black, fill=green!30, thick, minimum size=7mm]
\tikzstyle{switch} = [rectangle, draw=black, fill=blue!30, thick, minimum size=7mm]
\tikzstyle{cache} = [rectangle, draw=black, fill=purple!30, thick, minimum size=7mm]
\tikzstyle{input} = [circle, draw=black, fill=blue!30, thick, minimum size=12mm]

\makeatletter
\def\acknowledge{\gdef\@thefnmark{}\@footnotetext}
\makeatother

\newcommand{\bra}[1]{\left({#1}\right)}
\newcommand{\sbra}[1]{\left[{#1}\right]}
\newcommand{\cbra}[1]{\left\{{#1}\right\}}

\newcommand{\approptoinn}[2]{\mathrel{\vcenter{
  \offinterlineskip\halign{\hfil$##$\cr
    #1\propto\cr\noalign{\kern2pt}#1\sim\cr\noalign{\kern-2pt}}}}}

\def\BibTeX{{\rm B\kern-.05em{\sc i\kern-.025em b}\kern-.08em
    T\kern-.1667em\lower.7ex\hbox{E}\kern-.125emX}}

\makeatletter
\DeclareRobustCommand\onedot{\futurelet\@let@token\@onedot}
\def\@onedot{\ifx\@let@token.\else.\null\fi\xspace}

\makeatother

\begin{document}

\title{Regime Learning for Differentiable Particle Filters}

\makeatletter
\newcommand{\linebreakand}{
  \end{@IEEEauthorhalign}
  \hfill\mbox{}\par
  \mbox{}\hfill\begin{@IEEEauthorhalign}
}
\makeatother

\author{\IEEEauthorblockN{John-Joseph Brady}
\IEEEauthorblockA{\textit{Computer Science Research Centre} \\
\textit{University of Surrey}\\
Guildford, United Kingdom \\
j.brady@surrey.ac.uk}
\and
\IEEEauthorblockN{Yuhui Luo}
\IEEEauthorblockA{\textit{Data Science Department} \\
\textit{National Physical Laboratory}\\
Teddington, United Kingdom \\
yuhui.luo@npl.co.uk}
\and
\IEEEauthorblockN{Wenwu Wang}
\IEEEauthorblockA{\textit{Centre for Vision, Speech, and Signal Processing} \\
\textit{University of Surrey}\\
Guildford, United Kingdom \\
w.wang@surrey.ac.uk}
\linebreakand
\IEEEauthorblockN{V\'{i}ctor Elvira}
\IEEEauthorblockA{\textit{School of Mathematics} \\
\textit{University of Edinburgh}\\
Edinburgh, United Kingdom \\
victor.elvira@ed.ac.uk}
\and
\IEEEauthorblockN{Yunpeng Li}
\IEEEauthorblockA{\textit{Computer Science Research Centre} \\
\textit{University of Surrey}\\
Guildford, United Kingdom \\
yunpeng.li@surrey.ac.uk}
}
\maketitle

\begin{abstract}
 Differentiable particle filters are an emerging class of models that combine sequential Monte Carlo techniques with the flexibility of neural networks to perform state space inference. This paper concerns the case where the system may switch between a finite set of state-space models, \emph{i.e.} regimes. No prior approaches effectively learn both the individual regimes and the switching process simultaneously. In this paper, we propose the neural network based regime learning differentiable particle filter (RLPF) to address this problem. We further design a training procedure for the RLPF and other related algorithms. We demonstrate competitive performance compared to the previous state-of-the-art algorithms on a pair of numerical experiments.
\end{abstract}


\section{Introduction}
\acknowledge{The authors acknowledge funding from the UK Department for Science, Innovation and Technology through the 2024 National Measurement System programme. JJ. Brady is supported through the National Physical Laboratory and University of Surrey partnership via an Engineering and Physical Sciences Research Council studentship. The work of V. E. is supported by ARL/ARO under grant W911NF-22-1-0235 and the University of Edinburgh – Rice University Strategic Collaboration Award.}

Particle filters, first introduced in \cite{BPF}, are a class of Monte Carlo sampling algorithms that sequentially update the posterior distribution of the unobserved state of a dynamical system upon receipt of noisy measurements. Systems of this structure are known as state-space models (SSMs). Particle filters have found applications in target tracking \cite{Target-tracking}, robot localisation \cite{DPF, soft-resample}, and financial product risk analysis \cite{SVM}. Classical particle filtering algorithms require prior knowledge of the functional form of the SSM. Differentiable particle filters (DPFs) \cite{DPF, NF-filter} represent a recent effort to reduce the required prior knowledge by parameterising part of the model with flexible neural networks.

We consider the problem that the system of interest may vary dynamically between a discrete set of candidate state-space models or regimes. In \cite{MMPF}, the authors introduced a particle filter where the regime choice is assumed to be a realisation of a Markov chain. In \cite{RSPF} the regime switching particle filter (RSPF) was introduced as a generalisation to settings where the regime choice can depend arbitrarily on its history. The regime switching differentiable bootstrap particle filter (RSDBPF) \cite{RSDBPF} learns a neural network parameterisation of the RSPF using the framework of differentiable particle filtering. However, the RSDBPF still requires that the probabilistic meta-model that determines regime choice, hereafter referred to as the ``switching dynamic", is known a priori.

 In the particle filtering literature, there have been past efforts to handle systems where the switching dynamic is unknown. One such example is the model averaging particle filter (MAPF) \cite{MAPF} where a separate particle filter is run for each regime. Computational effort is assigned per filter according to the posterior probability of each regime. Optionally, the filters are allowed to occasionally exchange particles; this strategy is poorly suited to settings where the model is allowed to change frequently. 

In this paper, we introduce the regime learning particle filter (RLPF), which uses neural networks to parameterise the switching dynamic. Our contributions are two-fold:

\begin{enumerate}
    \item We develop a neural network parameterisation of regime switching systems, capable of learning the switching dynamic.
    \item We propose a novel algorithm to train differentiable particle filters with a Markovian marginal component, that we demonstrate empirically to improve accuracy.
\end{enumerate}


The structure of the paper is as follows: in Section \ref{sect:problem} we formally set up the problem the paper addresses; in Section \ref{sect:preliminaries} we review the required background knowledge. Section \ref{sect:RLPF} introduces the proposed regime learning particle filter algorithm, with Sections \ref{sect:redefine} and \ref{sect:LearnRS} introducing a novel parameterisation and Section \ref{sect:training} devoted to a proposed training strategy. In Section \ref{sect:experiments} we demonstrate competitive performance, with respect to prior state-of-the-art approaches, on the test environments adopted in previous works. We conclude this paper in Section \ref{sect:conclusions}.

\section{Problem Formulation}
\label{sect:problem}
A state-space model describes a system of two components, an unobserved discrete time Markov process $\cbra{\mathbf{x}_t}$ and its noisy observations $\cbra{\mathbf{y}_t}$. In our problem we allow the system to randomly jump, at any time, between a number of different SSMs, indexed by $\cbra{k_t}$. We impose the assumption that the choice of model, $\cbra{k_t}$, is independent of $\cbra{\mathbf{x}_t, \mathbf{y}_t}$, but can depend arbitrarily on its history. We illustrate this system graphically in Fig. \ref{fig:general-regime}, and represent it algebraically as:
\begin{equation}
\begin{gathered}
        k_{0} \sim K^{\theta}_{0}\bra{k_{0}} \, ,\\
        \mathbf{x}_{0} \sim M^{\theta}_{0}\bra{\mathbf{x}_0|k_{0}} \, ,\\
        k_{t\geq 1} \sim K^{\theta}\bra{k_{t}|k_{0:t-1}} \, ,\\
        \mathbf{x}_{t\geq 1} \sim M^{\theta}\bra{\mathbf{x}_{t}|\mathbf{x}_{t-1}, k_{t}} \, ,\\
        \mathbf{y}_{t} \sim G^{\theta}\bra{\mathbf{y}_{t}| \mathbf{x}_{t}, k_{t}} \, . \\
        \label{set-up}
\end{gathered}
\end{equation}
\begin{figure}
\centering
\begin{tikzpicture}
    \node(y0) [obs] {$\mathbf{y}_{0}$};
    \node(y1) [obs, right= 0.5cm of y0] {$\mathbf{y}_{1}$};
    \node(y2) [obs, right= 0.5cm of y1] {$\mathbf{y}_{2}$};
    \node(y3) [obs, right= 0.5cm of y2] {$\mathbf{y}_{3}$};
    \node(x0) [state, below= 0.5cm of y0] {$\mathbf{x}_{0}$};
    \node(x1) [state, right= 0.5cm of x0] {$\mathbf{x}_{1}$};
    \node(x2) [state, right= 0.5cm of x1] {$\mathbf{x}_{2}$};
    \node(x3) [state, right= 0.5cm of x2] {$\mathbf{x}_{3}$};
    \node(xdots) [state, right= 0.5cm of x3] {$\cdots$};
    \node(xT) [state, right= 0.5cm of xdots] {$\mathbf{x}_{T}$};
    \node(yT) [obs, above= 0.5cm of xT] {$\mathbf{y}_{T}$};
    \node(k0) [switch, below= 0.5cm of x0] {$k_{0}$};
    \node(k1) [switch, below= 0.5cm of x1] {$k_{1}$};
    \node(k2) [switch, below= 0.5cm of x2] {$k_{2}$};
    \node(k3) [switch, below= 0.5cm of x3] {$k_{3}$};
    \node(kdots) [switch, below= 0.47cm of xdots] {$\cdots$};
    \node(kT) [switch, below= 0.47cm of xT] {$k_{T}$};

    \draw[->, thick] (x0.north) -- (y0.south);
    \draw[->, thick] (x1.north) -- (y1.south);
    \draw[->, thick] (x2.north) -- (y2.south);
    \draw[->, thick] (x3.north) -- (y3.south);
    \draw[->, thick] (xT.north) -- (yT.south);
    \draw[->, thick] (k0.north) -- (x0.south);
    \draw[->, thick] (k1.north) -- (x1.south);
    \draw[->, thick] (k2.north) -- (x2.south);
    \draw[->, thick] (k3.north) -- (x3.south);
    \draw[->, thick] (kT.north) -- (xT.south);
    \draw[->, thick] (x0.east) -- (x1.west);
    \draw[->, thick] (x1.east) -- (x2.west);
    \draw[->, thick] (x2.east) -- (x3.west);
    \draw[->, thick] (x3.east) -- (xdots.west);
    \draw[->, thick] (xdots.east) -- (xT.west);
    \draw[->, thick] (k0.east) -- (k1.west);
    \draw[->, thick] (k1.east) -- (k2.west);
    \draw[->, thick] (k2.east) -- (k3.west);
    \draw[->, thick] (k3.east) -- (kdots.west);
    \draw[->, thick] (kdots.east) -- (kT.west);
    \draw[->, thick] (k0.south) to[out = -90, in = -90, distance = 0.5cm] (k1.south);
    \draw[->, thick] (k0.south) to[out = -90, in = -90, distance = 0.6cm] (k2.south);
    \draw[->, thick] (k0.south) to[out = -90, in = -90, distance = 0.7cm] (k3.south);
    \draw[->, thick] (k0.south) to[out = -90, in = -90, distance = 0.8cm] (kT.south);
    \draw[->, thick] (k1.south) to[out = -90, in = -90, distance = 0.5cm] (k2.south);
    \draw[->, thick] (k1.south) to[out = -90, in = -90, distance = 0.6cm] (k3.south);
    \draw[->, thick] (k1.south) to[out = -90, in = -90, distance = 0.7cm] (kT.south);
    \draw[->, thick] (k2.south) to[out = -90, in = -90, distance = 0.5cm] (k3.south);
    \draw[->, thick] (k2.south) to[out = -90, in = -90, distance = 0.6cm] (kT.south);
    \draw[->, thick] (k3.south) to[out = -90, in = -90, distance = 0.5cm] (kT.south);
    \draw[->, thick] (k0.north west) to[out = 180, in = 180, distance = 0.5cm] (y0.west);
    \draw[->, thick] (k1.north west) to[out = 180, in = 180, distance = 0.5cm] (y1.west);
    \draw[->, thick] (k2.north west) to[out = 180, in = 180, distance = 0.5cm] (y2.west);
    \draw[->, thick] (k3.north west) to[out = 180, in = 180, distance = 0.5cm] (y3.west);
    \draw[->, thick] (kT.north west) to[out = 180, in = 180, distance = 0.5cm] (yT.west);
\end{tikzpicture}
        \caption{Bayesian network representation of the general regime switching model.}
        \label{fig:general-regime}
\end{figure}
    
    Throughout this paper we will refer to $t$ as the ``time-index"; $\theta$ as the ``model parameters"; $k_{t}$, which may only take integers in the set $\mathcal{K} = \cbra{0, \dots, K}$, as the ``model index"; $\mathbf{x}_{t}$ as the ``latent state"; and $\mathbf{y}_{t}$ as the ``observations". The model components shall be referred to as follows: $K^{\theta}_{0},K^{\theta}$ as the ``switching dynamic"; $M^{\theta}_{0}, M^{\theta}$ as the ``dynamic models"; and $G^{\theta}$ as the ``observation models". For the sake of simplicity and to maintain clarity in our discussion, we allow the overloading of notations and use $M_0^{\theta}(\mathbf{x}_0)$, $M^{\theta}(\mathbf{x}_t|\mathbf{x}_{t-1})$ and $G^{\theta}(\mathbf{y}_t|\mathbf{x}_t)$ to represent the prior, dynamic model and measurement model, respectively, when there is a single regime, \emph{i.e.} the system is a vanilla SSM. We assume that during training we have access to the ground truth state, $\cbra{\mathbf{x}_t}$, but not the model indices, $\cbra{k_t}$.

    We desire to find an accurate estimator of $\mathbf{x}_{t}$ given $\mathbf{y}_{0:t}$, we propose to do this via a particle filtering estimate of $\mathbb{E}^{\theta}\sbra{\mathbf{x}_{t}|\mathbf{y}_{0:t}}$, denoted by $\hat{\mathbf{x}}_{t}$.
    
\section{Preliminaries}
\label{sect:preliminaries}
\subsection{Regime switching particle filtering}
\label{sect:RSPF}
    We first introduce a generic particle filtering framework. Classical particle filtering algorithms provide inference on SSMs, \emph{i.e.} our problem (Eq. \eqref{set-up}) if there were only one regime. 
    
    Given an SSM, a particle filter is a procedure to sequentially obtain an importance sample from its filtering distribution, $P\bra{\mathbf{x}_{t}| \mathbf{y}_{0:t}}$. A tutorial on particle filtering can be found in \cite{DoucetTutorial}. The full algorithm is illustrated in pseudo-code in Algorithm \ref{alg:PF}.

\begin{algorithm}[h]
\caption{Generic Particle Filter. All operations indexed by $i$ should be repeated for all $i \in \cbra{1, \dots, N}$.}
\label{alg:PF}
\begin{algorithmic}[1]
\REQUIRE \hspace*{0.3em} prior $M_{0}$ 
\hspace*{6.4em} dynamic model $M$\\
\hspace*{1.8em} proposal prior $Q_{0}$
\hspace*{2.9em} proposal $Q$ \\ 
\hspace*{1.8em} observation model $G$  
\hspace*{1.7em} time length $T$ \\
\hspace*{1.8em} particle count $N$ 
\hspace*{3.5em} observations $\mathbf{y}_{0:T}$\\
\ENSURE  particle locations $\tilde{\mathbf{x}}^{0:N}_{0:T}$
\hspace*{1em} particle weights $w^{0:N}_{0:T}$ \\
\hspace*{1.8em} normalised weights $\Bar{w}^{0:N}_{0:T}$
\STATE Sample $\tilde{\mathbf{x}}^{i}_{0} \sim Q_{0}\bra{\tilde{\mathbf{x}}^{i}_{0}}$;
\STATE $w^{i}_{0} \leftarrow \frac{M_{0}\bra{\tilde{\mathbf{x}}^{i}_{0}}}{Q_{0}\bra{\tilde{\mathbf{x}}^{i}_{0}}} G\bra{\mathbf{y}_{0}| \tilde{\mathbf{x}}^{i}_{0}}$; \\
\STATE $\bar{w}^{i}_{0} \leftarrow \frac{w^{i}_{0}}{\sum^{N}_{n}w^{n}_{0}}$;
\vspace{1pt}
\FOR{$t = 1$ to $T$}
    \STATE Set the resampled indices and weights, $A^{i}_{t}$, $\tilde{w}^{i}_{t}$, according to the chosen resampling scheme; \\
    \STATE Sample $\tilde{\mathbf{x}}^{i}_{t} \sim Q \bra{\tilde{\mathbf{x}}^{i}_{t}| \tilde{\mathbf{x}}^{A^{i}_{t}}_{t-1}}$; \\
    
    \STATE $w^{i}_{t} \leftarrow \tilde{w}^{i}_{t} \frac{M\bra{\tilde{\mathbf{x}}^{i}_{t}| \tilde{\mathbf{x}}^{A^{i}_{t}}_{t-1}}}{Q\bra{\tilde{\mathbf{x}}^{i}_{t}| \tilde{\mathbf{x}}^{A^{i}_{t}}_{t-1}}} G\bra{\mathbf{y}_{t}| \tilde{\mathbf{x}}^{i}_{t}}$;
    \STATE $\bar{w}^{i}_{t} \leftarrow \frac{w^{i}_{t}}{\sum^{N}_{n}w^{n}_{t}}$;
\ENDFOR
\RETURN $\tilde{\mathbf{x}}^{1:N}_{0:T}, \, w^{1:N}_{0:T}, \, \Bar{w}^{1:N}_{0,T}$.
\end{algorithmic}
\end{algorithm}

    Several quantities can be estimated from the particle approximation of the filtering distribution; of interest to us are  the approximation of the filtering mean, $\mathbb{E}\sbra{\mathbf{x}_t| \mathbf{y}_{0:t}}$:
    \begin{equation}
         \hat{\mathbf{x}}_{t} =  
        \sum^{N}_{i=1} \tilde{\mathbf{x}}^{i}_{t}\Bar{w}^{i}_{t},
        \label{filter-mean}
    \end{equation}
    where $N$ is the total number of particles; and the observation-likelihood:
    \begin{equation}
        \hat{p}\bra{\mathbf{y}_{0:t}} = \prod^{t}_{s=0}\frac{1}{N}\sum^{N}_{i=1}w^{i}_{t}.
        \label{likelihood}
    \end{equation}
    The mean, $\hat{\mathbf{x}}_{t}$, relies on auto-normalised importance weights and so is biased in general \cite{SMCbook}, but $\hat{p}\bra{\mathbf{y}_{0:t}}$ is unbiased \cite{UnbiasedLikelihood}. The derivation of both equations can be found in \cite{SMCbook}.
    
    If one is able to express a problem as an SSM then they can apply particle filtering (Alg. \ref{alg:PF}). For example, the regime switching problem (Eq. \eqref{set-up}) can be reformulated as an SSM by taking the latent state to be $\cbra{\mathbf{x}_{t}, k_{0:t}}$, and the observations as $\cbra{\mathbf{y}_{t}}$. This is the strategy the RSPF \cite{RSPF} uses to approximate the joint posterior $P\bra{\mathbf{x}_{t}, k_{t}| \mathbf{y}_{0:t}}$.

\subsection{Differentiable particle filtering}
\label{sect:DPFs}
Differentiable particle filters (DPFs) \cite{DPF, NF-filter} propose to learn an optimal parameter set via stochastic gradient descent (SGD) on some target loss function, $\mathcal{L}^{\theta}$. The adoption of SGD, rather than the traditionally preferred expectation maximisation (EM) algorithm \cite{SMC-opt}, allows the flexibility in model required to use a neural network parameterisation.

In DPFs the gradient of the loss is calculated by back-propagating through the filtering process. This raises an issue for the sampling steps; sampling a continuous distribution in a differentiable way is straightforward: choose the sampling output to be a differentiable, deterministic function of its input and some random variable that is independent of any values we want to pass gradient through -- this is known as ``the reparameterisation trick''. Unfortunately, in the case that the target distribution is discrete, as it is during resampling, no such differentiable function exists in general. An alternative approach is to sample from some prior distribution over the classes and perform importance sampling.

In practice, although the true prior on the resampling indices is uniform, a uniform proposal can be too far from the target distribution to produce effective importance samples. So, at the cost of biased gradients, we adopt a proposal that is a mixture of the target distribution, with some probability $\alpha$; and a uniform distribution, with probability $1-\alpha$. This is known as ``soft-resampling" \cite{soft-resample}. More recent unbiased resampling schemes, such as optimal transport-based resampling \cite{OT-resample}, do not apply as they require the latent state to be continuous. Further detail on DPFs can be found in \cite{DPF-Review}.

In supervised settings, defined as having access to both the ground truth latent state $\mathbf{x}_{0:T}$ as well as the observations $\mathbf{y}_{0:T}$ in training, it is common to directly use the target objective as the training loss. In the non-regime-switching analog to our problem, the appropriate loss function is the mean squared error between the estimated states and the ground truth, as previously used in \cite{DPF, soft-resample}:
\begin{equation}
    \mathcal{L}_{\text{MSE}}\bra{\hat{\mathbf{x}}_{0:T}, \mathbf{x}_{0:T}} = \frac{1}{T+1}\sum^{T}_{t=0}\lVert \hat{\mathbf{x}}_{t} - \mathbf{x}_{t} \rVert^{2}_{2} \, .
    \label{MSE}
\end{equation}
One can also consider the unsupervised problem, where only the observations are available during both training and testing. In this case it is common to use the usual variational inference approach of minimising an evidence lower bound \cite{naesseth2018variational, le2018auto-encoding, maddison2017variational}:
\begin{equation}
    \mathcal{L}_{\text{ELBO}}\bra{\mathbf{y}_{0:T}} =  -\log\bra{\hat{p}\bra{\mathbf{y}_{0:T}}} \, .
    \label{ELBO}
\end{equation}
Note that, whilst the estimator of the likelihood $\hat{p}\bra{\mathbf{y}_{0:T}}$ is unbiased, its back-propagated gradient is not. $\hat{\mathbf{x}}_{t}$ and $\hat{p}\bra{\mathbf{y}_{0:T}}$ are defined in Eqs. \eqref{filter-mean} and \eqref{likelihood} respectively.
    
 The regime switching differentiable particle filter (RSDBPF) was proposed in \cite{RSDBPF} to address the regime-switching filtering problem \eqref{set-up} where the switching dynamic is assumed to be known but the other model components are not. It can be thought of a RSPF where the parameters are learned by gradient descent. The RSDBPF does not employ soft-resampling, 
 and instead zeros out the contribution to the gradient of the particles due their ancestors at every time-step, improving gradient variance at the cost of bias. This is found empirically to improve performance in some cases \cite{diff-resample}.

 \section{The regime-learning particle filter}
 \label{sect:RLPF}
 In this section we propose the regime learning particle filter (RLPF). We first, in Section \ref{sect:redefine}, introduce an equivalent redefinition of the problem, Eq. \eqref{set-up}, that naturally leads to a structure which we paramaterise in Section \ref{sect:LearnRS}. Finally, we introduce a novel training strategy in Section \ref{sect:training}.
 \subsection{Redefining the model}
 \label{sect:redefine}
 We find it instructive, both from an implementation and an understanding perspective, to propose the following reformulation of the regime-switching dynamic in \eqref{set-up}:

 \begin{equation}
    \begin{gathered}
            k_{0} \sim K^{\theta}_{0}\bra{k_{0}} \, ,\\
            k_{t\geq 1} \sim K^{\theta}\bra{k_{t}|\mathbf{r}_{t-1}} \, ,\\
            \mathbf{r}_{t\geq 0} = R^{\theta}\bra{k_{t}, \mathbf{r}_{t-1}} \, .
            \label{our-set-up}
    \end{gathered}
    \end{equation}

\begin{figure}
\centering
\begin{tikzpicture}
    \node(y0) [obs] {$\mathbf{y}_{0}$};
    \node(y1) [obs, right= 0.5cm of y0] {$\mathbf{y}_{1}$};
    \node(y2) [obs, right= 0.5cm of y1] {$\mathbf{y}_{2}$};
    \node(y3) [obs, right= 0.5cm of y2] {$\mathbf{y}_{3}$};
    \node(x0) [state, below= 0.5cm of y0] {$\mathbf{x}_{0}$};
    \node(x1) [state, right= 0.5cm of x0] {$\mathbf{x}_{1}$};
    \node(x2) [state, right= 0.5cm of x1] {$\mathbf{x}_{2}$};
    \node(x3) [state, right= 0.5cm of x2] {$\mathbf{x}_{3}$};
    \node(xdots) [state, right= 0.5cm of x3] {$\cdots$};
    \node(xT) [state, right= 0.5cm of xdots] {$\mathbf{x}_{T}$};
    \node(yT) [obs, above= 0.5cm of xT] {$\mathbf{y}_{T}$};
    \node(k0) [switch, below= 0.5cm of x0] {$k_{0}$};
    \node(k1) [switch, below= 0.5cm of x1] {$k_{1}$};
    \node(k2) [switch, below= 0.5cm of x2] {$k_{2}$};
    \node(k3) [switch, below= 0.5cm of x3] {$k_{3}$};
    \node(dots) [draw=black, fill=gray!40, thick, minimum size=0.7cm, below= 1cm of xdots] {$\cdots$};
    \node(kT) [switch, below= 0.47cm of xT] {$k_{T}$};
    \node(r0) [cache, below= 0.45cm of k0] {$\mathbf{r}_{0}$};
    \node(r1) [cache, below= 0.45cm of k1] {$\mathbf{r}_{1}$};
    \node(r2) [cache, below= 0.45cm of k2] {$\mathbf{r}_{2}$};
    \node(r3) [cache, below= 0.45cm of k3] {$\mathbf{r}_{3}$};
    \node(rT) [cache, below= 0.45cm of kT] {$\mathbf{r}_{T}$};

    \draw[->, thick] (x0.north) -- (y0.south);
    \draw[->, thick] (x1.north) -- (y1.south);
    \draw[->, thick] (x2.north) -- (y2.south);
    \draw[->, thick] (x3.north) -- (y3.south);
    \draw[->, thick] (xT.north) -- (yT.south);
    \draw[->, thick] (k0.north) -- (x0.south);
    \draw[->, thick] (k1.north) -- (x1.south);
    \draw[->, thick] (k2.north) -- (x2.south);
    \draw[->, thick] (k3.north) -- (x3.south);
    \draw[->, thick] (kT.north) -- (xT.south);
    \draw[->, thick] (x0.east) -- (x1.west);
    \draw[->, thick] (x1.east) -- (x2.west);
    \draw[->, thick] (x2.east) -- (x3.west);
    \draw[->, thick] (x3.east) -- (xdots.west);
    \draw[->, thick] (xdots.east) -- (xT.west);
    \draw[->, thick] (k0.east) -- (k1.west);
    \draw[->, thick] (k1.east) -- (k2.west);
    \draw[->, thick] (k2.east) -- (k3.west);
    \draw[->, thick] (k3.east) to[out = 0, in = 180] (dots.west);
    \draw[->, thick] (dots.east) to[out = 0, in = 180] (kT.west);
    \draw[->, thick] (r0.east) -- (r1.west);
    \draw[->, thick] (r1.east) -- (r2.west);
    \draw[->, thick] (r2.east) -- (r3.west);
    \draw[->, thick] (r3.east) to[out = 0, in = 180] (dots.west);
    \draw[->, thick] (dots.east) to[out = 0, in = 180] (rT.west);
    \draw[->, thick] (k0.south) -- (r0.north);
    \draw[->, thick] (k1.south) -- (r1.north);
    \draw[->, thick] (k2.south) -- (r2.north);
    \draw[->, thick] (k3.south) -- (r3.north);
    \draw[->, thick] (kT.south) -- (rT.north);
    \draw[->, thick] (r0.east) to[out = 0, in = 180] (k1.west);
    \draw[->, thick] (r1.east) to[out = 0, in = 180] (k2.west);
    \draw[->, thick] (r2.east) to[out = 0, in = 180] (k3.west);
    \draw[->, thick] (r0.east) to[out = 0, in = 180] (k1.west);
    \draw[->, thick] (k0.north west) to[out = 180, in = 180, distance = 0.5cm] (y0.west);
    \draw[->, thick] (k1.north west) to[out = 180, in = 180, distance = 0.5cm] (y1.west);
    \draw[->, thick] (k2.north west) to[out = 180, in = 180, distance = 0.5cm] (y2.west);
    \draw[->, thick] (k3.north west) to[out = 180, in = 180, distance = 0.5cm] (y3.west);
    \draw[->, thick] (kT.north west) to[out = 180, in = 180, distance = 0.5cm] (yT.west);

\end{tikzpicture}

        \caption{Bayesian network representation of the proposed modified regime switching model.}
        \label{fig:modified-regime}
\end{figure}
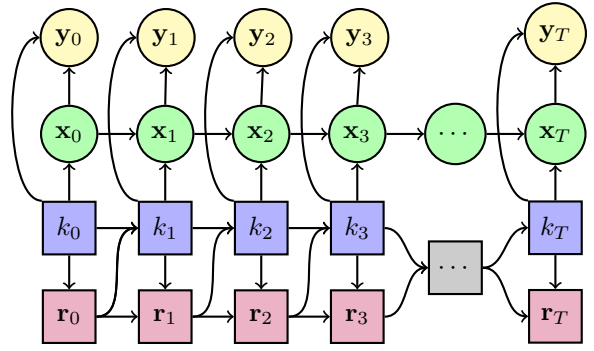
    
One can easily see that this formulation is equivalent to Eq. \eqref{set-up} by taking $R^{\theta}\bra{k_{t-1}, \mathbf{r}_{t-1}} = k_{t-1}\bigoplus \mathbf{r}_{t-1}$, where $\bigoplus$ denotes concatenation, but doing so requires an unbounded amount of memory. In practice, and due to our desire to parameterise the switching dynamic by neural networks, we apply the practical constraint $\mathbf{r}_{t} \subseteq \mathbb{R}^{d_{r}}$ for a constant dimensionality $d_{r}$. The problem of learning the switching dynamic then reduces to finding embedding functions $R^{\theta}$, and regime probability masses, $K^{\theta}$. $K^{\theta}_{0}$ can be represented as a learnable vector, and we define $\mathbf{r}_{-1}$ to be a vector of zeros, but alternatively it can be learned. We note that $\cbra{\mathbf{x}_{t}, k_{t}, \mathbf{r}_{t}}$ is a Markov process; it is clear that \eqref{our-set-up} is an SSM. So particle filtering (Alg. \ref{alg:PF}) can be use to estimate the joint posterior $P\bra{\mathbf{x}_{t}, k_{t}, \mathbf{r}_{t}| \mathbf{y}_{0:t}}$. We illustrate this formulation as a Bayesian network diagram in Fig. \ref{fig:modified-regime}.

\subsection{Parameterising the switching dynamic}
\label{sect:LearnRS}
    We propose a neural network parameterisation of the switching dynamic. It is well known that particle filters for which the late-time state depends strongly on the early-time state do not perform well, in fact, permitting strong dependence, one can construct filters that diverge in variance at any desired rate \cite{SMCbook}. Late-time particles are genealogically non-diverse at early time and so form poor samples of the early-time state; a phenomenon often referred to as ``path-degeneracy". For this reason, we take inspiration from the architecture of the long-short term memory (LSTM) unit \cite{LSTM} and build forget gates into our model as:
    \begin{equation}
    \begin{split}
        \mathbf{r}_{t} =& R^{\theta}\bra{\mathbf{k'}_{t}, \mathbf{r}_{t-1}} \\
        =& \sigma\bra{\Theta_{1} \mathbf{r}_{t - 1}} \odot \sigma \bra{\Theta_{2} \mathbf{k'}_{t}} \odot \mathbf{r}_{t -1} \\
        &+ \text{tanh}\bra{\Theta_{3}  \mathbf{k'}_{t}} \odot \sigma\bra{\Theta_{4}  \mathbf{k'}_{t}},
    \end{split}
        \label{R}
    \end{equation}
    where $\mathbf{k'}_{t}$ is the one-hot enconding of the model index; $\cbra{\Theta_{1:4}}$ are matrices of weights to be learned; $\odot$ denotes the Hadamard product. 
    \begin{equation}
    \begin{gathered}
        K'^{\theta}\bra{\mathbf{k'}_{t}|\mathbf{r}_{t-1}} = \lvert \Theta_{5}\text{tanh}\bra{\Theta_{6}\mathbf{r}_{t-1}}\rvert \cdot \mathbf{k'}_t, \\
        K^{\theta}\bra{\mathbf{k'}_{t}|\mathbf{r}_{t-1}} = \frac{K'^{\theta}_{t}\bra{\mathbf{k'}_{t}|\mathbf{r}_{t-1}}}{\sum_{c \in \mathcal{K}} K'^{\theta}\bra{\mathbf{c'}|\mathbf{r}_{t-1}}},
    \end{gathered}
    \label{Q}
    \end{equation}
    where $\cdot$ denotes a dot product, and $\Theta_{5}, \Theta_{6}$ are additional learnable weight matrices. We represent the model graphically in Fig. \ref{fig:RSNET}.
    
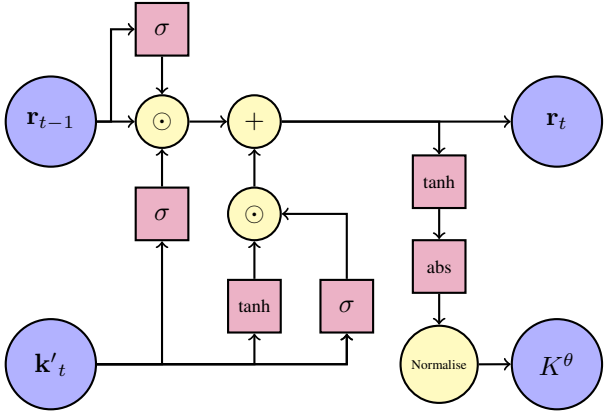
\begin{figure}
\centering
\begin{tikzpicture}
    \node (inr) [input] {$\mathbf{r}_{t-1}$};
    \node (ink) [input, below = 2cm of inr] {$\mathbf{k'}_{t}$};
    \node (prod1) [obs, right =  0.5cm of inr] {$\odot$};
    \node (sig1) [cache, above = 0.5cm of prod1] {$\sigma$};
    \node (sig2) [cache, below = 0.5cm of prod1] {$\sigma$};
    \node (plus) [obs, right = 0.5cm of prod1] {$+$};
    \node (prod2) [obs, below = 0.5cm of plus] {$\odot$};
    \node (tanh1) [cache, below = 0.5cm of prod2] {\scriptsize tanh};
    \node (sig3) [cache, right = 0.5cm of tanh1] {$\sigma$};
    \node (abs) [cache, above right = -0.2 and 0.5cm of sig3] {\scriptsize abs};
    \node (norm) [obs, below = 0.41cm of abs] {\tiny Normalise};
    \node (tanh2) [cache, above = 0.4cm of abs] {\scriptsize tanh};
    \node (outr) [input, right = 5.5cm of inr] {$\mathbf{r}_{t}$}; 
    \node (outk) [input, below = 2cm of outr] {$K^{\theta}$}; 

    \draw[->, thick] (inr.east) -- (prod1.west);
    \draw[->, thick] (sig1.south) -- (prod1.north);
    \draw[->, thick] (inr.east) -- (0.8, 0) |- (sig1.west);
    \draw[->, thick] (prod1.east) -- (plus.west);
    \draw[->, thick] (plus.east) -- (outr.west);
    \draw[->, thick] (prod2.north) -- (plus.south);
    \draw[->, thick] (ink.east) -| (sig2.south);
    \draw[->, thick] (sig2.north) -- (prod1.south);
    \draw[->, thick] (ink.east) -| (tanh1.south);
    \draw[->, thick] (tanh1.north) -- (prod2.south);
    \draw[->, thick] (ink.east) -| (sig3.south);
    \draw[->, thick] (ink.east) -| (sig3.south);
    \draw[->, thick] (sig3.north) |- (prod2.east);
    \draw[->, thick] (plus.east) -| (tanh2.north);
    \draw[->, thick] (tanh2.south) -- (abs.north);
    \draw[->, thick] (abs.south) -- (norm.north);
    \draw[->, thick] (norm.east) -- (outk.west);
\end{tikzpicture}
    \caption{Graphical representation of our proposed switching dynamic. Blue nodes are input/outputs, purple nodes are fully connected network layers with the specified activation, and yellow nodes are non-learned functions. The switching probability mass, $K^{\theta}\bra{k_{t+1}|\mathbf{r}_{t}}$, is the value at the $k_{t+1}^{\text{th}}$ index of the model output $K^{\theta}$.}
    \label{fig:RSNET}
\end{figure}

    As with resampling, $K^{\theta}\bra{k_{t}|\mathbf{r}_{t-1}}$ is a discrete distribution; in order to back-propagate gradients through the network we use importance sampling to select the model indices.

\subsection{Training the RLPF}
\label{sect:training}
We adopt a novel training strategy. Instead of treating the problem as a supervised regression on $\hat{\mathbf{x}}_{0:T}$ with the MSE-loss $\mathcal{L}_{\text{MSE}}\bra{\hat{\mathbf{x}}_{0:T}, \mathbf{x}_{0:T}}$ (Eq. \eqref{MSE}) as in previous work \cite{RSDBPF}, we introduce an additional $\mathcal{L}_{\text{ELBO}}\bra{\cbra{\mathbf{x}_{0:T}, \mathbf{y}_{0:T}}}$ (Eq. \eqref{ELBO}) unsupervised loss term in the training objective:
\begin{equation}
\begin{split}\mathcal{L}^{\theta}_{\text{RLPF}}\bra{\hat{\mathbf{x}}_{0:T}, \mathbf{x}_{0:T}, \mathbf{y}_{0:T}} = \mathcal{L}_{\text{ELBO}}\bra{\cbra{\mathbf{x}_{0:T}, \mathbf{y}_{0:T}}} + \\ \lambda \mathcal{L}_{\text{MSE}}\bra{\hat{\mathbf{x}}_{0:T}, \mathbf{x}_{0:T}} \, ,
\end{split}
\label{our-loss}
\end{equation}
where $\lambda \in \mathbb{R}^{+}$ is a hyper-parameter.

For the supervised term, we run a particle filter on an SSM defined as in Section \ref{sect:redefine}, with the state as $\cbra{\mathbf{x}_{t}, k_{t}, \mathbf{r}_{t}}$ and the observation as $\mathbf{y}_{t}$. For the unsupervised term, we treat all the information we have access to in training as observed, \emph{i.e.} take $\cbra{\mathbf{x}_{t}, \mathbf{y}_{t}}$ to be the observations. However, in doing so the observations depend on their history not only through the state, so is not, by our definition, an SSM. However, treating the past observations as constant, it is easy to show that the usual particle filtering algorithm remains correct. We refer to this strategy as a ``marginal filter" as it runs a particle filter only on the marginal process that is the switching dynamic.
The intuition behind this strategy is that training on the ELBO loss maximises the data likelihood to bring the learned model closer to the true model; and the MSE loss serves to guide the training towards a lower validation objective.

The validation objective is the MSE calculated by running a filter with the $\theta$ learned by optimising the combined loss. Furthermore, there is no need for the algorithm to be differentiable during evaluation, so there we use the non-differentiable systematic resampling \cite{systematic-resample} and the target distribution, $K^{\theta}$, to propose the model indices.


  We present the full RLPF procedure in Algorithm \ref{alg:RLPF}.\footnote{Our implementation, that was used to generate all results reported in this paper, can be found at: \url{https://github.com/John-JoB/Regime_Switching}}

\begin{algorithm}[t]
\caption{Regime Learning Particle Filter. The $\bullet$ operator denotes multiplication of probability densities, and $\beta$ is the learning rate according to optimiser choice.}
\label{alg:RLPF}
\begin{algorithmic}[1]
\REQUIRE priors $M^{\theta}_{0}$
\hspace*{6.1em} dynamic models $M^{\theta}$\\
\hspace*{1.3em} regime prior $K^{\theta}_{0}$
\hspace*{3.5em} switching dynamic $K^{\theta}$\\
\hspace*{1.3em} observation models $G^{\theta}$
\hspace*{0.7em} encoding functions $R^{\theta}$ \\
\hspace*{1.3em} time length $T$
\hspace*{4.5em} particle count $N$ \\
\hspace{1.3em} loss coefficient $\lambda$
\hspace*{3.2em} observations $\mathbf{y}_{0:T}$ \\
\hspace*{1.3em} ground truth $\mathbf{x}_{0:T}$
\ENSURE  model parameters $\theta$
\WHILE{$\theta$ not converged}
\STATE Run a particle filter (Algorithm \ref{alg:PF}) to obtain $\tilde{\mathbf{x}}^{1:N}_{0:T}, \Bar{w}^{1:N}_{0:T}$ with inputs: \\
        $M_{0} \leftarrow K^{\theta}_{0} \bullet M^{\theta}_{0} \bullet R^{\theta}$
        \hspace*{0em} $Q_{0} \leftarrow \text{Uniform}\bra{k \in \mathcal{K}} \bullet M^{\theta}_{0} \bullet R^{\theta}$ \\
        $M \leftarrow K^{\theta} \bullet M^{\theta} \bullet R^{\theta}$
        \hspace*{0em} $Q \leftarrow \text{Uniform}\bra{k \in \mathcal{K}} \bullet M^{\theta} \bullet R^{\theta}$ \\
        $G \leftarrow G^{\theta}$
        \hspace*{4.7em} $T \leftarrow T$ \\
        $N \leftarrow N$
        \hspace*{4.7em} $\mathbf{y}_{0:T} \leftarrow \mathbf{y}_{0:T}$
\vspace{2pt}
\STATE $\hat{\mathbf{x}}_{0:T} \leftarrow \sum^{N}_{n=1}\tilde{\mathbf{x}}^{n}_{0:T}\Bar{w}^{n}_{0:T}$, Eq. \eqref{filter-mean}
\vspace{2pt}
\STATE $\mathcal{L}^{\theta}_{\text{MSE}} \leftarrow \frac{1}{T+1}\sum^{T}_{t=0}\lVert \hat{\mathbf{x}}_{t} - \mathbf{x}_{t} \rVert$, Eq. \eqref{MSE}
\vspace{2pt}
\STATE Run a particle filter (Algorithm \ref{alg:PF}) to obtain $w^{1:N}_{0:T}$ with inputs: \\
        $M_{0} \leftarrow K^{\theta}_{0} \bullet R^{\theta}$,
        \hspace*{1.6em} $Q_{0} \leftarrow \text{Uniform}\bra{k \in \mathcal{K}} \bullet R^{\theta}$ \\
        $M \leftarrow K^{\theta} \bullet R^{\theta}$
        \hspace*{2em} $Q \leftarrow \text{Uniform}\bra{k \in \mathcal{K}} \bullet R^{\theta}$ \\
        $G_{0} \leftarrow G^{\theta} \bullet M^{\theta}_{0}$
        \hspace*{1.6em} $G_{t\geq1} \leftarrow G^{\theta} \bullet M^{\theta}\bra{\cdot | \mathbf{x}_{t-1}}$\\
        $T \leftarrow T$
        \hspace*{5.4em}$N \leftarrow N$ \\
        $\mathbf{y}_{0:T} \leftarrow \cbra{\mathbf{y}_{0:T}, \mathbf{x}_{0:T}}$
\vspace{2pt}
\STATE $\mathcal{L}^{\theta}_{\text{ELBO}}\leftarrow -  \sum^{T}_{t=0}\log\bra{\frac{1}{N}\sum^{N}_{n=1} w^{n}_{t}}$, Eqs. \eqref{likelihood}, \eqref{ELBO}
\STATE $\mathcal{L}^{\theta}_{\text{RLPF}} \leftarrow \mathcal{L}^{\theta}_{\text{ELBO}} + \lambda\mathcal{L}^{\theta}_{\text{MSE}}$, Eq. \eqref{our-loss}
\STATE $\theta \leftarrow \theta + \beta \nabla_{\theta} \mathcal{L}^{\theta}_{\text{RLPF}}$
\ENDWHILE
\RETURN $\theta$
\end{algorithmic}
\end{algorithm}

\section{Experiments}
\label{sect:experiments}
We repeat the experiment set-up adopted in \cite{RSPF} and \cite{RSDBPF}, where $x_{t}, y_{t} \in \mathbb{R}$ and, for each of the eight regimes, the dynamic model and observation model are linear Gaussian and non-linear Gaussian respectively. 

\subsection{Dynamic and observation models}
\small
\begin{equation}
    \begin{gathered}
        M_0\bra{x_0} = \mathcal{U}\bra{-0.5, 0.5}\,\,, \\
        M\bra{x_{t}| x_{t-1}, k_{t}} = \mathcal{N}\bra{\mu_{M} \bra{x_{t-1}, k_{t}}, \sigma^{2}}\,\,, \\
        G\bra{y_{t}| x_{t}, k_{t}} = \mathcal{N}\bra{\mu_{G} \bra{x_{t}, k_{t}}, \sigma^{2}}\,\,, \\
        \mu_{M} \bra{x_{t-1}} = a_{k_{t}} x_{t-1} + b_{k_{t}}\,\,, \\
        \mu_{G} \bra{x_{\tau}} = a_{k_{t}} \sqrt{\lvert x_{t} \rvert} + b_{k_{t}}\,\,, \\
        \sbra{a_{1}, \dots, a_{8}} = \sbra{-0.1, -0.3, -0.5, -0.9, 0.1, 0.3, 0.5, 0.9}\,\,, \\
        \sbra{b_{1}, \dots, b_{8}} = \sbra{0, -2, 2, -4, 0, 2, -2, 4}\,\,, \\
        \sigma^{2} = 0.1\,\,,
    \end{gathered}
    \label{experiment-setup}
\end{equation}
\normalsize
where $M_{0}\bra{x_0}, M\bra{x_{t}| x_{t-1}, k_{t}}, G\bra{y_{t}| x_{t}, k{t}}$ are the true models, and we have defined eight discrete regimes. Interestingly, $G\bra{y_{t}|x_{t}, k_{t}}$ is bimodal, so accurate estimation of the state at time $t$ requires using information over time-steps. For example, Regimes $1$ and $5$ lead to identical distributions on the observations despite their disparate posteriors. So, knowledge of the switching dynamic is crucial in accurately estimating the full posterior of the switching system.

\subsection{Switching dynamic}
We test our algorithm on two switching dynamics: a Markov switching system and a P\'{o}lya-Urn distribution. For both dynamics, $K^{\theta}_{0}\bra{k_{0}}$ is uniform over $\mathcal{K}.$ The Markov switching dynamic at non-zero time can be expressed as:
\begin{equation}
    \begin{gathered}
    K\bra{k_{t}|k_{0:t-1}} = \bra{\mathbf{k'}_{t}}^{T} B \mathbf{k'}_{t-1}\,\,, \\
    B = 
    \begin{pmatrix}
        0.8 & 0.15 & \rho & \dots & \rho \\
        \rho & 0.8 & 0.15 & \dots & \rho \\
        \vdots & & \ddots & & \vdots \\
        \rho & \rho & \dots & 0.8 & 0.15 \\
        0.15 & \rho & \dots & \rho & 0.8 
    \end{pmatrix}\,\,,\\
    \rho = \frac{1}{120}\,\,,
    \end{gathered}
    \label{markov}
\end{equation}
where, as before, $\mathbf{k'}_{t}$ is the one-hot encoding of $k_{t}$. And the P\'{o}lya-Urn distribution:
\begin{equation}
    K\bra{k_{t}|k_{0:t-1}} = \frac{ 1 + \sum^{t}_{s=0}\mathds{1}\bra{k_{s} = k_{t}}}{8 + \sum^{8}_{c=1}\sum^{t}_{s=0}\mathds{1}\bra{k_{s} = c}}\,\,.
    \label{polya}
\end{equation}

\subsection{Baseline Models}
The point estimation of states can be considered as a sequence-to-sequence supervised learning task; we desire to learn sequence $x_{0:T}$ given the sequence $y_{0:T}$. As such, we can apply any state-of-the-art sequence-to-sequence learning algorithms. We compare with the baseline models of a unidirectional LSTM \cite{LSTM}, and a decoder-only transformer model\cite{Transformer}. The recurrent structure of an LSTM is more alike the underlying SSMs; however, as transformer models have been receiving growing attention, we include it for completeness.

The LSTM and Transformer baselines cannot provide an interpretable statistical output, as the proposed model does. So, we introduce two particle filtering baselines. The first is to run a DPF where the latent state is taken to be $x_{t}$ only. This strategy implicitly assumes that there is no regime switching so we expect poor performance in the examined settings. The second is a modification of the model averaging particle filter (MAPF) \cite{MAPF} to make it differentiable, which we name ``the model averaging differentiable particle filter" (MADPF). The MAPF was proposed to solve the inverse to the problem addressed by the RSDBPF; it requires knowledge of the individual models but not the switching dynamic. The MADPF is a MAPF trained by gradient descent, analogous to how the RSDBPF is derived from the RSPF.

We include two oracle approaches: the RSDBPF which simulates from the true switching dynamic but learns the individual regimes; and the RSPF which corresponds to a particle filter run on the true model.

The novel training strategy proposed in Section \ref{sect:training} and the parameterisation of the switching dynamic, in Sections \ref{sect:redefine} and \ref{sect:LearnRS}, can be applied independently. So we include two ablation models, an RLPF trained on solely the MSE loss, and an RSDBPF trained on our proposed loss, equation \eqref{our-loss}.

\subsection{Experiment settings}
We generate 2,000 trajectories from the model (Eqs. \eqref{experiment-setup} with \eqref{markov} or \eqref{polya}) and use 1,000 for training, $500$ for validation, and $500$ for testing. All algorithms are initialised and trained from scratch for 20 repeats on a different generation of the dataset. During training and validation, particle filters are run with 200 particles; this is increased to 2,000 for testing. The parameter set that obtained the best validation loss is passed to testing. We use the Adam optimiser with weight decay \cite{AdamW}.

Throughout the experiments, we parameterise each $\mu_{G}$ and $\mu_{M}$ with a two-layer fully connected neural network with hyperbolic tangent activation on the hidden layer. We set the dimensionality of $\mathbf{r}_{t}$ to the number of models. All other hyper-parameters are chosen by grid search, on a per-experiment basis. Soft resampling is only used for the marginal filter. 
When using $\mathcal{L}^\theta_{\text{RLPF}}$, we impose a prior that the dynamic and observation model variances are less than 1; or else the ELBO is lowered rapidly in the first few iterations by raising these variances, instead of approximating the true model.

It was shown in \cite{RSPF} that, for these experiments, the RSPF performs equally well whether the model indices are proposed uniformly or from the target distribution ($K^{\theta}\bra{k_{t}}$). So, we use a uniform proposal during training. However, in a case where there are a large number of very unlikely regimes, a uniform proposal would lead to highly inefficient sampling. In such a case, introducing bias-inducing soft-sampling, as we did in section \ref{sect:DPFs} for resampling, might improve performance.
\subsection{Results}

We present the results for both experiments in Table \ref{tab:results} including baselines and ablation results, and a comparison of the computation costs under the P\'{o}lya-Urn experiment in Table \ref{tab:time} for the non-oracle methods. The Markov times were similar so are excluded for brevity. For the RLPF and the RSDBPF the qualification -$\lambda$ or -MSE denotes whether the experiment used the proposed training algorithm (Section \ref{sect:training}), or the classical MSE only approach. All timing experiments were performed using an NVIDA RTX 4090 GPU.

For both experiments, the proposed RLPF-$\lambda$ leads to the smallest mean squared errors among non-oracle algorithms. Moreover, being a particle filter, it has a statistical interpretation; with it one can evaluate properties such as uncertainties, the data likelihood, and make predictions about future state or data. The LSTM is the next best in the reported error metrics, but it does not offer the statistical properties seen in particle filters. We also note that for both the RSDBPF and the RLPF, the variants trained using the proposed training objective achieved lower MSEs and variances than the respective pure-MSE approaches.

Aside from the Transformer, all algorithms considered scale linearly in time with trajectory length. However, the relatively simple architecture of the LSTM and better ability to take advantage of GPU parallelism give it a significant speed advantage, as shown in Table~\ref{tab:time}. Achieving better parallelism for the RLPF to bring its time cost closer to that of the DPF is left for future work.

\begin{table}[htbtp]
    \centering
    \caption{Filtering accuracy for the discussed algorithms. Reported values are the achieved mean squared filtering error, averaged across 20 independent training runs.}
    \begin{tabular}{||c|c|c||}
    \hline
        Algorithm & Markov MSE & P\'{o}lya MSE \\
    \hline
    \hline
    Transformer (baseline) &$1.579\pm0.169$&$1.508\pm0.112$\\
     \hline
     LSTM (baseline) &$0.713\pm0.114$&$0.655\pm0.027$\\
     \hline
     DBPF (baseline) &$4.689\pm4.689$&$2.395 \pm 0.723$\\
     \hline
     MADPF (baseline) &$2.688 \pm 0.607$&$1.987\pm0.171$\\
     \hline
     RLPF-MSE  (baseline) &$0.970\pm0.187$&$0.693\pm0.080$\\
     \hline
     RLPF-$\lambda$ (proposed) &$\mathbf{0.698\pm0.164}$&$\mathbf{0.613\pm0.072}$\\
     \hline
     \hline
     RSDBPF-$\lambda$ (partial oracle) &$0.802\pm0.128$&$0.590\pm0.064$\\
     \hline
     RSDBPF-MSE (partial oracle) &$1.153\pm0.217$&$0.709\pm0.122$\\
     \hline
     RSPF (oracle) &$0.274\pm0.019$&$0.413\pm0.012$\\
     \hline
    \end{tabular}
    \label{tab:results}
\end{table}

\begin{table}[htbtp]
    \centering
    \caption{Average computation times per training epoch (10 batches of 100 parallel filters of 200 particles each), and testing run (1 batch of 500 parallel filters of 2000 particles each) on the P\'{o}lya experiment.}
    \begin{tabular}{||c|c|c||}
    \hline
        Algorithm & Av. train epoch time (s) & Av. test time (s) \\
    \hline
    \hline
    Transformer (baseline) &$0.182$&$0.00310$\\
     \hline
     LSTM (baseline) &$\mathbf{0.0145}$&$\mathbf{0.000792}$\\
     \hline
     DBPF (baseline) &$0.553$&$0.243$\\
     \hline
     MADPF (baseline) &$18.6$&$3.73$\\
     \hline
     RLPF-MSE  (baseline) &$4.42$&$0.620$\\
     \hline
     RLPF-$\lambda$ (proposed) &$6.66$&$0.612$\\
     \hline
    \end{tabular}
    \label{tab:time}
\end{table}

\section{Conclusions}
\label{sect:conclusions}
In this paper we have proposed a novel particle filtering approach suited to situations where the model may switch, according to some unknown dynamic, between a set of candidate models whose forms are also unknown. We designed a learning strategy and demonstrated its superiority experimentally. Future directions include the incorporation of more advanced DPFs, more interpretable regime identification procedure, and more complex test scenarios including real world data.

\bibliographystyle{IEEEtran}
\bibliography{ref}
\end{document}